\documentclass[11pt]{article}

\usepackage{amsmath,amssymb}
\usepackage{graphicx}
\usepackage{float}
\usepackage{hyperref}
\usepackage{natbib}
\usepackage{setspace}
\usepackage{geometry}
\usepackage{subcaption}
\title{\textbf{Expert System for Bitcoin Forecasting: Integrating Global Liquidity via TimeXer Transformers}}

\date{}

\author{
Sravan Karthick T\thanks{Corresponding author. Email: \texttt{sravankt.cs20@rvce.edu.in}} \\
RV College of Engineering (RVCE), Bengaluru, India
\and
Rakshita A \\
RV College of Engineering (RVCE), Bengaluru, India
\and
Dr. Minal Moharir \\
RV College of Engineering (RVCE), Bengaluru, India
}

\begin{document}
\maketitle

\begin{abstract}
Bitcoin price forecasting is characterized by extreme volatility and non-stationarity,
making it hard for univariate time-series prediction models over long-horizons. By integrating Global M2 Liquidity as an exogenous variable, aggregated from 18 major economies, we improve the predictions over long-horizons.

Using TimeXer, we compare a liquidity-conditioned univariate
forecasting model (TimeXer-Exog) against state-of-the-art univariate models, Long Short-Term Memory(LSTM), Neural Basis Expansion Analysis for interpretable Time-Series (N-BEATS), Patch Time-Series Transformer(PatchTST), and standard TimeXer. Experiments conducted
on daily Bitcoin price data from January 2020 to August 2025 demonstrate that
explicit macroeconomic conditioning significantly stabilizes long-horizon forecasts.

At a 70-day forecast horizon, the proposed TimeXer-Exog model achieves a Mean Squared Error (MSE) of $10.814$, scaled by $10^{7}$, which outperformed the univariate
TimeXer baseline by over 89\%. The results show that explicitly conditioning deep learning models on Global Liquidity as an exogenous variable gives substantial improvements in long-horizon Bitcoin price forecasting.
\end{abstract}


\onehalfspacing

\section{Introduction}
\label{introduction}

Early attempts to forecast Bitcoin prices were made using statistical tools, especially the ARIMA models \citep{Sagheer2025, Xu2025}. ARIMA is good at picking up short-term linear patterns, like the ones in a relatively calm equity index over a few weeks. However, bitcoin does not behave like that. Its price is noisy and very volatile, which statistical models and tools do not capture. ARIMA requires stationarity and linear structure, which is exactly what bitcoin lacks. Large rallies, sudden crashes, or regime shifts often get smoothed away \citep{Pratas2023, Mizdrakovic2024}. As a result, these models can perform well over a short horizon, their accuracy drops once the forecast horizon is about 60 or 90 days \citep{Mousa2025, Kareem2025}.

Researchers moved to RNN models, which were designed to handle sequences. Long Short-Term Memory (LSTM) and Gated Recurrent Unit (GRU) models were successful in learning the non-linear patterns from the data \citep{Bagheri2025, Xu2025}. They can, for example, react differently to a slow upward trend than to a sudden spike triggered by a regulatory headline. These models are not perfect. Because they process data step by step, they scale poorly and struggle when dependencies stretch far back in time. Capturing relationships across hundreds of days still remains hard, and unreliable \citep{Zhao2022, Zheng2025}.

More recently, attention has shifted toward Transformer-based models, time series forecasting moved to self-attention rather than strict sequence processing \citep{Nie2022, Zhao2022}. Instead of remembering the past one step at a time, these models can look across an entire window at once. PatchTST takes this idea further by breaking the time series into patches, small segments that preserve local structure while keeping computation manageable \citep{Nie2022}. For time series forecasting which require long look-back windows, Transformer models have repeatedly outperformed LSTM-based approaches, especially for long-horizon predictions \citep{Nie2022, Zheng2025}.

Long-term forecasting improves when models are allowed to look beyond price history itself. The TimeXer framework reflects this shift. It combines patch-wise attention for the main series with cross-attention for external variables, making it possible to integrate data that arrive at different frequencies and are not neatly aligned in time \citep{Wang2024}. This is closer to how real financial systems operate. Macroeconomic indicators do not update on a daily basis, yet markets tend to respond to them gradually and unevenly.

Global liquidity is often measured by M2 money supply, which has particularly attracted attention \citep{Gu2025}. Although the relationship is not perfectly stable, the logic is intuitive. According to empirical work using time-varying Granger causality, the influence of M2 on Bitcoin prices is not constant. It strengthens during expansionary regimes and weakens in other regimes \citep{Gu2025}. Co-integration analyses go further by pointing towards a long-run equilibrium relationship in which Bitcoin prices respond more than proportionally to liquidity growth \citep{Kokabian2025}.  It is important to note that these effects do not come into the picture suddenly. The impact of M2 unfolds with lags which corresponds to the idea that before reaching the crypto market, liquidity works its way through portfolios and risk appetite \citep{Kokabian2025}.

Our aim is to develop an \textbf{expert forecasting system} that transcends simple pattern recognition by integrating long-horizon modeling with explicit macroeconomic context. By embedding Global M2 Liquidity as an exogenous variable within the TimeXer framework, we enable the model to emulate the decision-making process of a human macro-strategist. Much like an expert analyst who weighs monetary policy shifts against technical price action, our system uses cross-attention mechanisms to dynamically attend to global liquidity cycles, thereby stabilizing predictions over longer horizons.

From a broader financial engineering perspective, this represents a meaningful evolution from univariate forecasting to \textbf{intelligent decision support}. Context is as critical as computation. While black-box predictors often fail during structural breaks, this approach incorporates domain-specific expert knowledge—specifically, the lead-lag relationship between money supply and risk assets. By moving from a univariate setup to a multivariate expert framework, the proposed approach delivers insights that are robust to the drift and structural changes common in digital asset markets, effectively functioning as an automated expert for capital allocation.

\section{Related Work}
\label{related_work}

\subsection{Bitcoin forecasting using ARIMA}

Bitcoin price predictions were conducted using ARIMA initially, largely because it was already a familiar tool in financial time series analysis \citep{Kareem2025, Xu2025}. For short-term movements, especially when prices drift without too much drama, ARIMA can be very reliable. One can imagine fitting it over a calm stretch of daily log returns and getting something that looks sensible for the next few days. The trouble starts when the market becomes volatile. Bitcoin prices are rarely stationary for too long, sharp jumps or sudden collapses violate the linear structure ARIMA assumes \citep{Sagheer2025, Pratas2023}. During volatile episodes, the model has a habit of smoothing away extreme moves, as if the crash or rally were just noise rather than the main event \citep{Kareem2025}. Forecasts also lose credibility quickly as the forecast horizon increases, which makes ARIMA a weak candidate for longer-term planning \citep{Mousa2025}.

\subsection{Bitcoin forecasting using CNN, RNN, and LSTM models}

As attention shifted toward more flexible approaches, deep learning models began to dominate the discussion.\citep{Bagheri2025}. Models such as RNNs and LSTMs performed particularly well and seemed to suit the task. They remembered patterns over longer periods of time and handled Bitcoin’s
nonlinear behavior better than classical models \citep{Hochreiter1997, Mardjo2024}. Researchers combined CNNs with LSTMs, letting convolutional layers pick up local price patterns, such as short bursts of momentum, before passing them to a recurrent structure \citep{Sagheer2025, Bagheri2025}. Several comparative analyses found that GRUs often matched or even outperformed LSTMs, while using fewer parameters and training more efficiently \citep{Kaur2025, Mohammadjafari2024, Nafisah2025}. However, these models have a common drawback. Since they process sequences step by step, they often become cumbersome when dependencies are from the past, which is often the case in crypto markets because they are shaped by long liquidity cycles.

\subsection{Bitcoin forecasting using Transformer-based models}

To tackle these limitations, Transformer models entered the scene \citep{Zheng2025}. They relied on self-attention rather than strict sequential processing and they examined longer windows of data in parallel. This makes it more efficient to capture distant dependencies without the trouble of training bottlenecks typical of RNNs. Time-series Transformers have delivered encouraging and positive results in cryptocurrency markets which have volatile settings \citep{Zheng2025}. In PatchTST, we divide the input series into patches. It can be considered as a practical compromise that preserves local structure while keeping attention costs under control \citep{Nie2022}. Results across a range of long-horizon tasks indicate that Transformer models have repeatedly outperformed LSTM baselines, producing forecasts that are not only more accurate but also less erratic over time \citep{Nie2022}. In conclusion, architecture alone does not explain everything, and context still matters.

\subsection{Incorporating exogenous variables in Bitcoin forecasting}

Relying solely on price data is not effective for long-term Bitcoin predictions. Researchers have experimented with on-chain activity, sentiment measures, and macroeconomic indicators to reflect broader market forces \citep{Mahfooz2024, Grubisic2025}. Global liquidity turned out to have a significant linkage \citep{Gu2025}. The connection is not mechanical, but there is evidence of a long-run equilibrium relationship between M2 and Bitcoin prices, with the influence becoming more pronounced during periods of monetary expansion \citep{Gu2025, Kokabian2025}. It is important to note that these effects tend to emerge with lags rather than instant reactions. Transformer-based models like TimeXer make it easier to incorporate such variables by allowing cross-attention between price dynamics and external drivers \citep{Wang2024}. This line of work points toward forecasting systems that treat Bitcoin not as an isolated series, but as part of a wider monetary environment, where global liquidity plays a role in shaping long-horizon price behavior.

\section{Methodology}

\subsection{The TimeXer Architecture}
 TimeXer was built to ease bringing external information into time series forecasting without disturbing the Transformer framework. TimeXer reuses the familiar Transformer backbone and adapts how inputs are represented and attended to. The result feels less like a radical redesign and more like a careful rethinking of how different data streams should talk to each other.

\begin{figure}[H]
    \centering
    \includegraphics[width=1.0\linewidth]{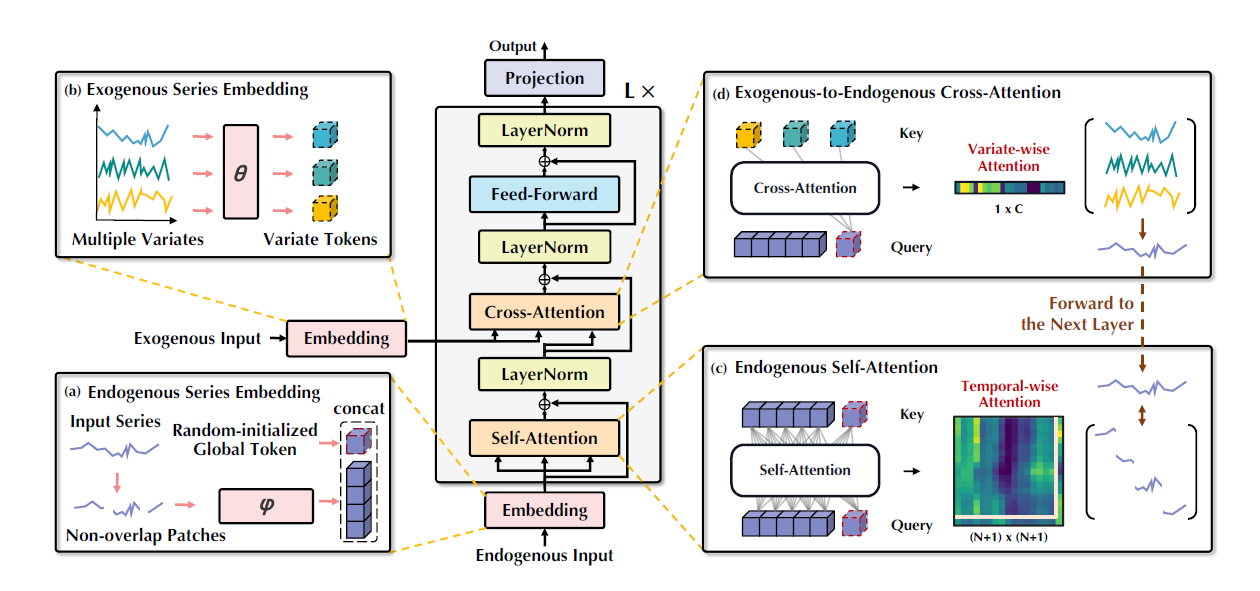}
    \caption{The TimeXer architecture showcasing the integration of endogenous and exogenous data streams through specialized embedding and attention mechanisms.}
    \label{fig:timexer_arch}
\end{figure}

\paragraph{Exogenous Input Integration and Cross-Attention.} 
TimeXer is composed of three token types to capture both internal temporal dynamics and external causal factors: patch-level temporal tokens and a learnable global token for the endogenous Bitcoin series, and series-wise variate tokens for exogenous variables.

\begin{enumerate}
    \item \textbf{Endogenous:} The primary Bitcoin price series (endogenous variable) is split into non-overlapping \textbf{patches} to finely capture temporal dependencies. The \textbf{learnable global token} for the endogenous series, acts as a macroscopic representation to aggregate information across the entire series.
    \item \textbf{Exogenous:} Global Liquidity is embedded as \textbf{series-wise variate tokens}. This representation handles practical irregularities which exist in real-world data, such as temporal misalignment, missing values, and mismatched frequencies.
\end{enumerate}

As shown in Figure \ref{fig:timexer_arch}, the model processes the input through two parallel pathways before combining them in the Transformer backbone. The \textbf{endogenous} (a) divides the Bitcoin price series into non-overlapping patches and adds a randomly initialized global token, forming the main temporal representation. In parallel, the \textbf{exogenous} (b) processes the Global Liquidity series separately, embedding it into variate-wise tokens that retain information about the external macroeconomic driver.

Both representations are then passed to the Transformer backbone. Here, \textbf{patch-wise self-attention} (c) captures temporal patterns within the Bitcoin series, while \textbf{variate-wise cross-attention} (d) uses the endogenous global token to selectively incorporate information from the exogenous variables. This design allows the model to learn Bitcoin’s internal dynamics while continuously adjusting to the broader macroeconomic signals from global liquidity.

\subsection{Baseline Models}
To benchmark performance, several established deep learning models were used for univariate forecasting:
\begin{itemize}
    \item \textbf{TimeXer (Endogenous Only):} This configuration serves as a control, testing the framework's native capability by relying solely on temporal patterns extracted from the Bitcoin price series.
    \item \textbf{PatchTST:} This Transformer variant utilizes \textbf{subseries-level patches} along with channel independence to enhance long-term forecasting accuracy \citep{Nie2022}.
    \item \textbf{LSTM:} A recurrent architecture designed to mitigate the vanishing gradient problem and capture \textbf{long-term dependencies} in sequential data \citep{Hochreiter1997, Mahfooz2024}.
    \item \textbf{N-BEATS:} A purely deep learning architecture consisting of a \textbf{very deep stack of fully-connected layers} linked by residual connections \citep{Oreshkin2019}.
\end{itemize}
\paragraph{Exclusion of Linear Baselines.}
Classical linear models, such as ARIMA and ARIMAX, were excluded from the final comparative analysis for the long-horizon tasks ($H=70$). These models rely on recursive (iterative) forecasting strategies to generate multi-step predictions. In high-volatility environments like cryptocurrency markets, recursive strategies suffer from severe \textit{error accumulation}, where slight deviations in early steps propagate and amplify over the forecast horizon \citep{Hyndman2018}. Furthermore, the stationarity transformations required for ARIMA to function (e.g., differencing) often result in a loss of the long-term trend information that is critical for the specific task of tracking global liquidity cycles. Consequently, we focus our benchmarking exclusively on deep learning architectures capable of direct multi-step forecasting.

\subsection{Bitcoin and Global Liquidity Relationship}
The integration of Global M2 Liquidity is justified by its documented long-run influence on Bitcoin pricing. A cointegration analysis established a strong relationship, characterized by a long-run elasticity estimate of \textbf{2.65}, meaning a 1\% increase in the M2 is associated with a 2.65\% increase in the price of Bitcoin \citep{Kokabian2025}. The stability of this long-run relationship is validated by a statistically significant Error Correction Term (ECT), $\lambda' = -0.12$ \citep{Kokabian2025}. The magnitude of this coefficient demonstrates that approximately \textbf{12\% of any deviation} from the long-run equilibrium is corrected monthly.

\begin{figure}[H]
    \centering
    \includegraphics[width=0.9\linewidth]{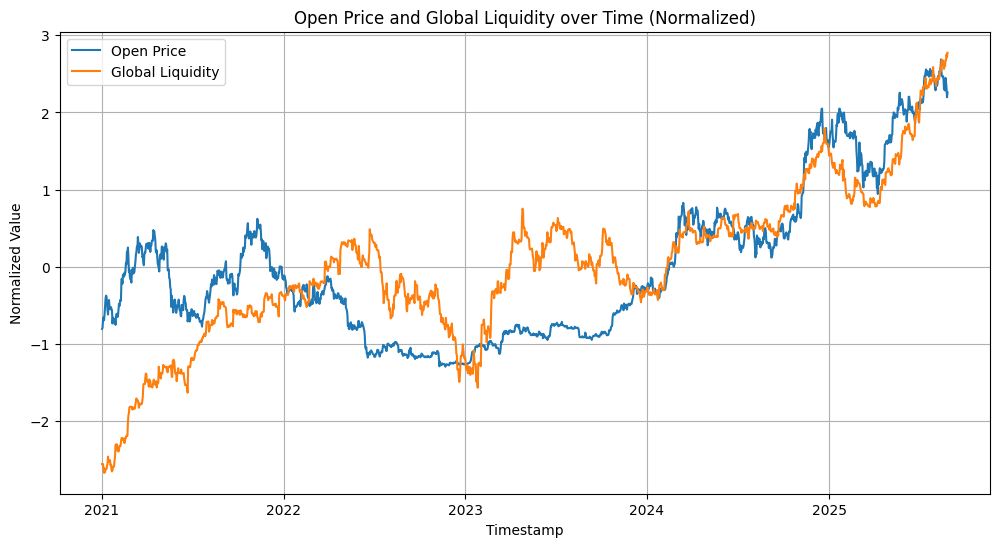}
    \caption{Bitcoin open price v/s Global Liquidity with a 12-week lead (both are standard normalized)}
    \label{fig:liquidity_correlation}
\end{figure}

 Figure \ref{fig:liquidity_correlation} highlights a clear lead-lag relationship between Global Liquidity and Bitcoin's price action. Shifting the M2 liquidity data back by 12 weeks, the alignment of the major cyclical peaks and troughs can be clearly observed. This observation suggests that changes in the global liquidity act as a leading indicator, with a transmission delay of approximately one fiscal quarter before the liquidity is fully reflected in Bitcoin's price. TimeXer leverages this 12-week lead as a predictive "look-ahead" window, thus improving the predictions.

\subsection{Statistical Evaluation: Model Confidence Set (MCS)}

To evaluate the predictive performance of the proposed \textit{TimeXer-Exog} model against the set of benchmark models ($\mathcal{M}_0 = \{ \text{TimeXer, PatchTST, LSTM, N-BEATS} \}$), we use MCS \citealp{Hansen2011MCS}.

Unlike traditional tests (e.g., the Diebold-Mariano test \citep{Diebold1995}), which can suffer from the multiple comparisons problem and data snooping bias when many models are evaluated \citep{White2000, Sullivan1999}, the MCS procedure determines a subset of models $\mathcal{M}^* \subset \mathcal{M}_0$ that contains the "true best" models with a specified confidence level $(1-\alpha)$. We iteratively eliminate models found to be significantly inferior until the null hypothesis of Equal Predictive Ability (EPA) cannot be rejected for the remaining set \citep{Hansen2011MCS}.

Since we use long-horizon forecasting (Horizon $h=70$), the forecast errors are expected to exhibit serial correlation up to lag $h-1$. We implemented the \textbf{Block Bootstrap} method \citep{Kunsch1989} within the MCS framework to account for this dependency structure and prevent spurious significance. In order to preserve the internal correlation structure of the error series during resampling \citep{Politis2004}, we set the block size equal to the forecast horizon ($h$) and the significance level to $\alpha = 0.10$. This means models with a p-value $< 0.10$ are excluded from the Superior Set Model (SSM).


\section{Data and Experimental Setup}
\label{data_experimental_setup}

\subsection{Data Description}

The endogenous dataset consists of daily Bitcoin (BTC) closing prices spanning from 
\textbf{January 1, 2020, to January 21, 2025}. This period has multiple market 
regimes, including monetary expansion phases, heightened volatility episodes, and 
prolonged drawdowns. This provides the perfect environment for evaluating long-horizon 
forecasting performance.

To ensure strict out-of-sample evaluation, the dataset was partitioned into training, validation, and test sets. The final 120 daily observations, from January 22, 2025, to May 21, 2025, were reserved exclusively for testing. The preceding 120 days, from September 24, 2024, to January 21, 2025, were used as the validation set for hyperparameter tuning and early stopping. All models were trained on the historical data prior to this validation window (January 1, 2020, to September 23, 2024).

\begin{table}[H]
\caption{Temporal Partitioning of the Dataset.}
\centering
\begin{tabular}{lllc}
\hline
\textbf{Dataset Split} & \textbf{Start Date} & \textbf{End Date} & \textbf{Length (Days)} \\ \hline
Training & January 1, 2020 & September 23, 2024 & 1,728 \\
Validation & September 24, 2024 & January 21, 2025 & 120 \\
Test & January 22, 2025 & May 21, 2025 & 120 \\ \hline
\textbf{Total} & \textbf{January 1, 2020} & \textbf{May 21, 2025} & \textbf{1,968} \\ \hline
\end{tabular}
\label{tab:data_partition}
\end{table}

\paragraph{Experimental reference date and temporal alignment.}
\textbf{27 August 2025} was fixed as the reference date for this study, 
which represents the latest point in time at which both Bitcoin price data and Global 
Liquidity information were assumed to be fully observable. To prevent any form of 
look-ahead bias, all exogenous inputs were strictly restricted to values available on or 
before this date.

Since Global Liquidity was incorporated using a dense lag structure spanning 
\textbf{1 to 98 days}, the effective sample end date is determined by the maximum lag 
applied. Specifically, applying a 98-day lag implies that the most recent usable Global 
Liquidity observation corresponds to \textbf{21 May 2025} (i.e., 98 days prior to 
27 August 2025). Consequently, after lag construction and temporal alignment, the final 
row of the modeling dataset ends on \textbf{21 May 2025}, even though the experiment 
itself is conceptually anchored to late August 2025.

This design ensures strict temporal causality: at every training and evaluation step, 
the model only has access to information that would have been available at the decision 
time, thereby eliminating forward-looking bias while preserving the full liquidity 
transmission window.

The exogenous variable employed in this study is \textbf{Global Liquidity}, constructed 
as an aggregate measure of broad money supply across a diversified set of major and 
emerging economies. Specifically, the indicator includes M2 from the following economies, using the closest available national definition where necessary: the United States, Euro Area, China, 
Japan, United Kingdom, Canada, Australia, India, South Korea, Brazil, Russia, 
Switzerland, Mexico, Indonesia, Turkey, Saudi Arabia, Argentina, and South Africa.

All national money supply series were converted into U.S. dollars using corresponding 
daily foreign exchange rates prior to aggregation. The resulting Global Liquidity 
measure therefore reflects worldwide monetary expansion in a unified currency framework, 
rather than being dominated by any single economy.

Empirical evidence suggests that changes in global liquidity affect Bitcoin prices with 
a delayed transmission. To account for this effect, the aggregated Global Liquidity 
series was shifted forward by \textbf{12 weeks}, reflecting the estimated average market 
adjustment period through which liquidity conditions propagate into digital asset 
valuations.

\subsection{Exogenous Input Construction}

Although the 12-week shift captures the dominant lead–lag relationship between Global 
Liquidity and Bitcoin prices, the forecasting framework does not rely on a single fixed 
lag. Importantly, the forecast horizon $H$ denotes the length of the prediction window 
and \textbf{does not correspond to the liquidity lag itself}.

Rather than imposing a predetermined lag structure, a \textbf{dense lag configuration} 
was constructed to allow the model to learn the full liquidity transmission profile 
endogenously. Specifically, multiple lagged versions of the Global Liquidity series were 
generated at the following offsets:
\[
1, 7, 14, 21, 28, 35, 42, 49, 56, 63, 70, 77, 84, 91, \text{ and } 98 \text{ days}.
\]

These lagged features are denoted as:
\[
\text{global\_lag}_1, \text{global\_lag}_7, \ldots, \text{global\_lag}_{98}.
\]

This dense lag structure enables the TimeXer architecture to infer both short- and 
long-delay effects of global liquidity on Bitcoin prices without explicitly specifying 
a functional form. Through variate-wise cross-attention, the model dynamically assigns 
importance to different liquidity lags depending on the forecast horizon, allowing it 
to adaptively capture regime-dependent transmission dynamics.

\subsection{Training Setup}

No manual normalization or scaling was applied during preprocessing. Instead, 
standardization was handled internally by each model through built-in normalization 
mechanisms such as \textbf{Reversible Instance Normalization (RevIN)} \citep{Kim2022} used in PatchTST. 
This approach ensures that normalization statistics are computed independently for each 
sample, thereby eliminating potential information leakage across time.

All models were trained using the \textbf{Mean Squared Error (MSE)} loss function:
\[
\text{MSE} = \frac{1}{N} \sum_{i=1}^{N} (y_i - \hat{y}_i)^2,
\]
where $y_i$ and $\hat{y}_i$ denote the observed and predicted Bitcoin prices, 
respectively. MSE was selected because it penalizes large forecast errors heavily, 
which is particularly important in cryptocurrency markets which are filled with price 
swings.

\subsection{Hyperparameter Optimization}
\label{hyperparameter_optimization}

To confirm that the model configurations were not chosen arbitrarily, a systematic hyperparameter search using \textbf{Optuna} was performed, an automated optimization framework employing the \textbf{Tree-structured Parzen Estimator (TPE)} algorithm \citep{akiba2019optuna}. The TPE algorithm was selected for its efficiency in sampling high-dimensional search spaces as compared to traditional grid or random search methods.

Minimizing the MSE on the validation set was the primary optimization objective (September 24, 2024, to January 21, 2025). For the Transformer-based models (TimeXer and PatchTST), the search space was defined to explore various model depths, attention capacities, and temporal receptive fields. The specific ranges and discrete values used during the optimization process are listed in Table \ref{tab:search_space}.

\begin{table}[H]
\caption{Hyperparameter Search Space for Transformer-based Models.}
\centering
\begin{tabular}{ll}
\hline
\textbf{Hyperparameter} & \textbf{Search Space / Range} \\ \hline
Input Size (Lookback) & $\{128, 256, 512\}$ \\
Patch Length & $\{32, 64, 96, 128\}$ \\
Stride & $\{8, 16, 32\}$ \\
Hidden Size ($d_{model}$) & $\{64, 128, 256\}$ \\
Encoding Layers & $\{4, 8, 12, 16\}$ \\
Attention Heads & $\{4, 8\}$ \\
Dropout & $[0.2, 0.4]$ (Continuous) \\
Learning Rate & $[10^{-6}, 10^{-4}]$ (Log-uniform) \\ \hline
\end{tabular}
\label{tab:search_space}
\end{table}

The optimization trials revealed that deeper architectures with larger patch sizes yielded the most stable long-term forecasts for Bitcoin. Specifically, a patch length of 96 with a stride of 8 provided the optimal balance between capturing local volatility and maintaining computational efficiency. Table \ref{tab:final_hyperparams} presents the final optimal configurations selected for each model based on the best validation performance.

\begin{table}[H]
\caption{Final Optimized Hyperparameters for All Models.}
\centering
\resizebox{\textwidth}{!}{%
\begin{tabular}{l|ccccc}
\hline
\textbf{Hyperparameter} & \textbf{TimeXer-Exog} & \textbf{TimeXer} & \textbf{PatchTST} & \textbf{LSTM} & \textbf{N-BEATS} \\ \hline
Input Size (Lookback) & 256 & 256 & 256 & 256 & 256 \\
Patch Length & 96 & 96 & 96 & - & - \\
Stride & 8 & 8 & 8 & - & - \\
Hidden Size ($d_{model}$) & 128 & 128 & 128 & 256 (Enc) / 128 (Dec) & 512 (MLP) \\
Layers & 16 & 16 & 6 & 3 (Enc) / 2 (Dec) & 3 Blocks \\
Attention Heads & 8 & 8 & 8 & - & - \\
Dropout & \textbf{0.30} & 0.25 & 0.20 - 0.30 & 0.15 & 0.0 \\
FFN Dimension ($d_{ff}$) & 512 & 512 & 512 & - & - \\
\hline
\end{tabular}%
}
\label{tab:final_hyperparams}
\end{table}

A consistent training configuration was employed specifically to ensure stable convergence across all architectures. All models were optimized using the \textbf{Adam optimizer} \citep{kingma2014adam}, which is well-suited for non-stationary financial data due to its adaptive learning rate capabilities. We utilized a fixed \textbf{batch size of 128}, chosen to balance memory efficiency with gradient stability. Considering the high volatility of the Bitcoin price series, this batch size provided sufficiently robust gradient estimates to prevent erratic weight updates while maintaining high computational throughput on the GPU. Standard scaling (z-score normalization) was applied to inputs to further aid the optimizer in navigating the loss landscape.

\subsection{Implementation Using the NeuralForecast Library}
All deep learning models were implemented using the \texttt{NeuralForecast} library, which 
provides a state-of-the-art framework for reproducing and training deep learning forecasting models \citep{Olivares2022}. 
The \texttt{NeuralForecast} framework was used to train and evaluate all the models (TimeXer, PatchTST, LSTM, 
and N-BEATS). 

The library has native support for future exogenous variables, thus we could easily integrate 
lagged Global Liquidity features into the TimeXer-Exog model. Model training 
was conducted using horizon-specific fitting with early stopping based on validation loss.
Rolling-origin forecasting was employed to assess true out-of-sample performance. 
This standardized implementation ensured fair comparison across all the models and minimized 
confounding effects arising from differences in training or evaluation procedures.

\subsection{Evaluation Metrics and Forecast Horizons}

Forecasts were generated for fixed horizons of
\[
H = \{7, 14, 21, 28, 35, 42, 49, 56, 63, 70\} \text{ days}.
\]

Performance was evaluated on the held-out out-of-sample test window using MSE as the evaluation metric. For numerical clarity and 
comparability across horizons, all the MSE values are scaled by a factor of 
$10^{7}$, with lower values indicating better predictions.

This evaluation method focuses on long-term stability and consistent price direction, rather than trying to match short-term fluctuations. This matches the study’s goal of understanding how macro-financial factors like global liquidity help improve long-term Bitcoin price predictions.

\section{Results and Discussion}

\subsection{Quantitative Performance}

\begin{table}[H]
\caption{Comparative MSE ($\times 10^{7}$) Across Forecast Horizons (Lower is Better). 
Best performance per horizon is shown in bold.}
\centering
\begin{tabular}{|c|ccccc|}
\hline
Horizon (H) & TimeXer-Exog & TimeXer & LSTM & NBEATS & PatchTST \\ \hline
70 & \textbf{10.814} & 101.952 & 29.120 & 77.276 & 147.566 \\
63 & \textbf{12.363} & 85.436 & 318.285 & 77.960 & 128.121 \\
56 & \textbf{5.363} & 62.699 & 26.653 & 23.399 & 45.092 \\
49 & \textbf{6.912} & 20.000 & 12.100 & 14.263 & 33.567 \\
35 & \textbf{4.580} & 9.447 & 22.720 & 17.933 & 22.971 \\
28 & \textbf{5.588} & 7.333 & 9.060 & 16.923 & 15.374 \\
21 & 4.087 & \textbf{3.585} & 26.804 & 14.579 & 13.263 \\
14 & \textbf{4.288} & 5.395 & 21.284 & 12.181 & 5.781 \\
7  & 3.160 & 4.351 & 2.810 & 4.843 & \textbf{2.028} \\
\hline
\end{tabular}
\label{tab:mse_results}
\end{table}

For the longest horizon, 70 days, the difference is clear. TimeXer-Exog outperforms the univariate TimeXer baseline by more than 89 percent.

Looking at Table~\ref{tab:mse_results}, a clear pattern emerges as the forecast horizon increases. In short windows, 7 to 14 days, the models that rely only on price history perform well. PatchTST outperforms TimeXer-Exog. Over one or two weeks, Bitcoin prices are mostly influenced by recent trends and short-term trading activity that are already reflected in the price data.

TimeXer-Exog begins to dominate after the one month mark. From 28 days onward, TimeXer-Exog has the lowest MSE and has great stability. There are less drifts and there seems to be a clearer sense of direction. Univariate models such as LSTM and NBEATS, performed poorly as the forecast horizon increased. Without outside information, their predictions either become flat or drift aimlessly, therefore showing the limits of using only price data to forecast.

The strong performance of TimeXer-Exog at longer horizons shows macro-financial is improving with the predictions. Providing lagged global M2 liquidity gives the model access to slow-moving forces that tend to lead market behavior rather than follow it. Using cross-attention, liquidity signals guide the prediction when price signals weaken, which seems to be the main reason for the improved predictions in long forecast horizons.

\subsection{Qualitative Forecast Analysis}

\begin{figure}[H]
\centering
\includegraphics[width=0.9\linewidth]{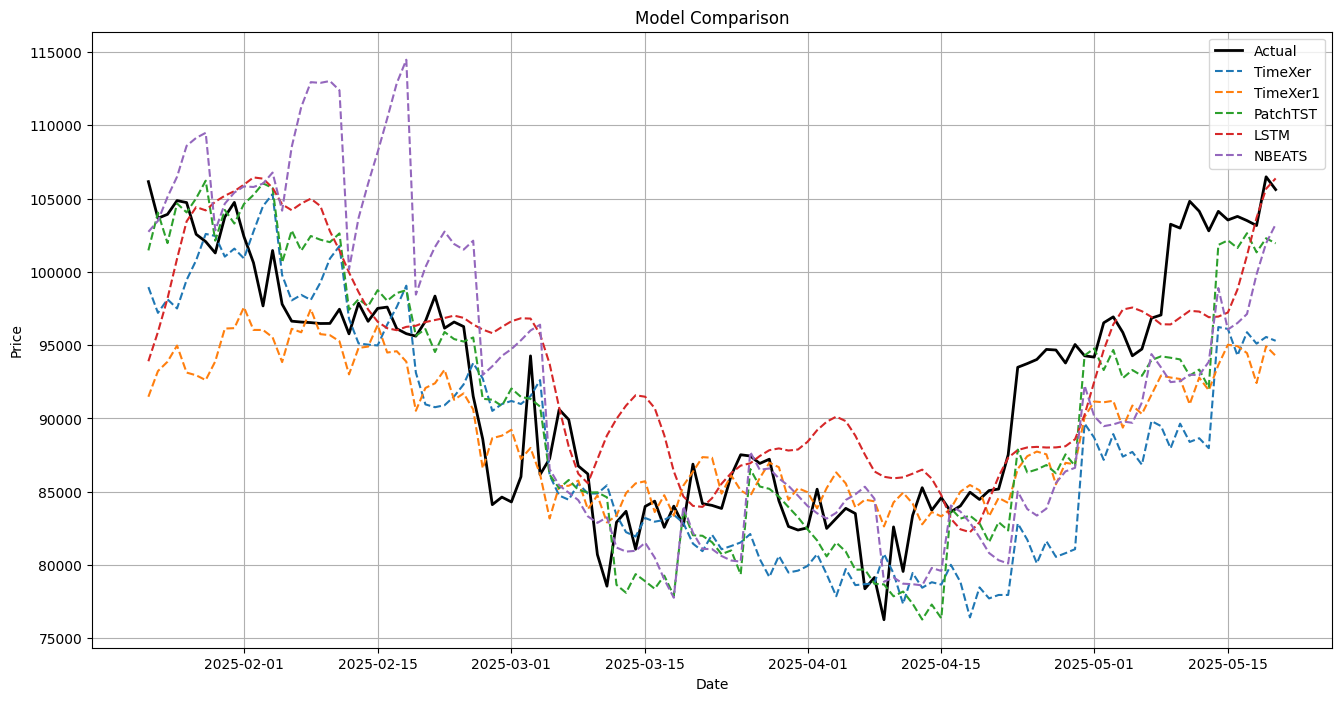}
\caption{Forecast comparison at horizon $H=7$ days.}
\label{fig:h7_forecast}
\end{figure}

\begin{figure}[H]
\centering
\includegraphics[width=0.9\linewidth]{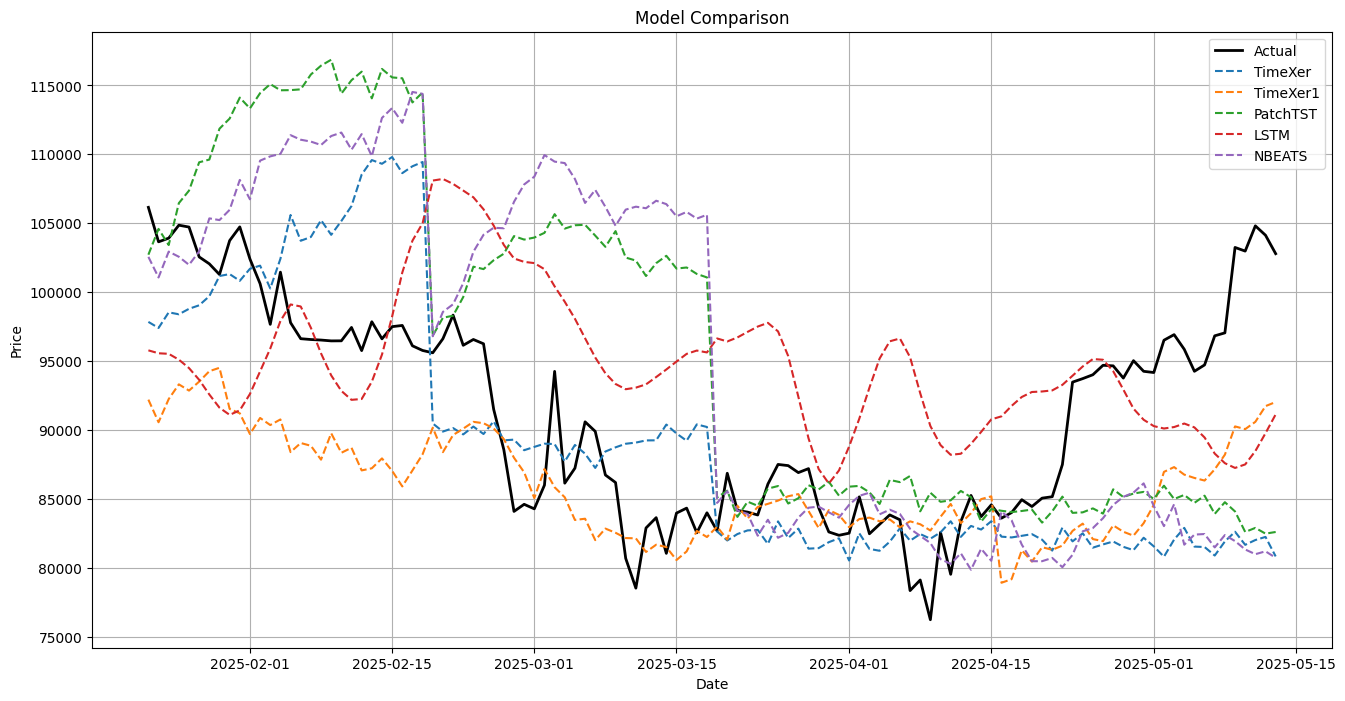}
\caption{Forecast comparison at horizon $H=28$ days.}
\label{fig:h28_forecast}
\end{figure}

\begin{figure}[H]
\centering
\includegraphics[width=0.9\linewidth]{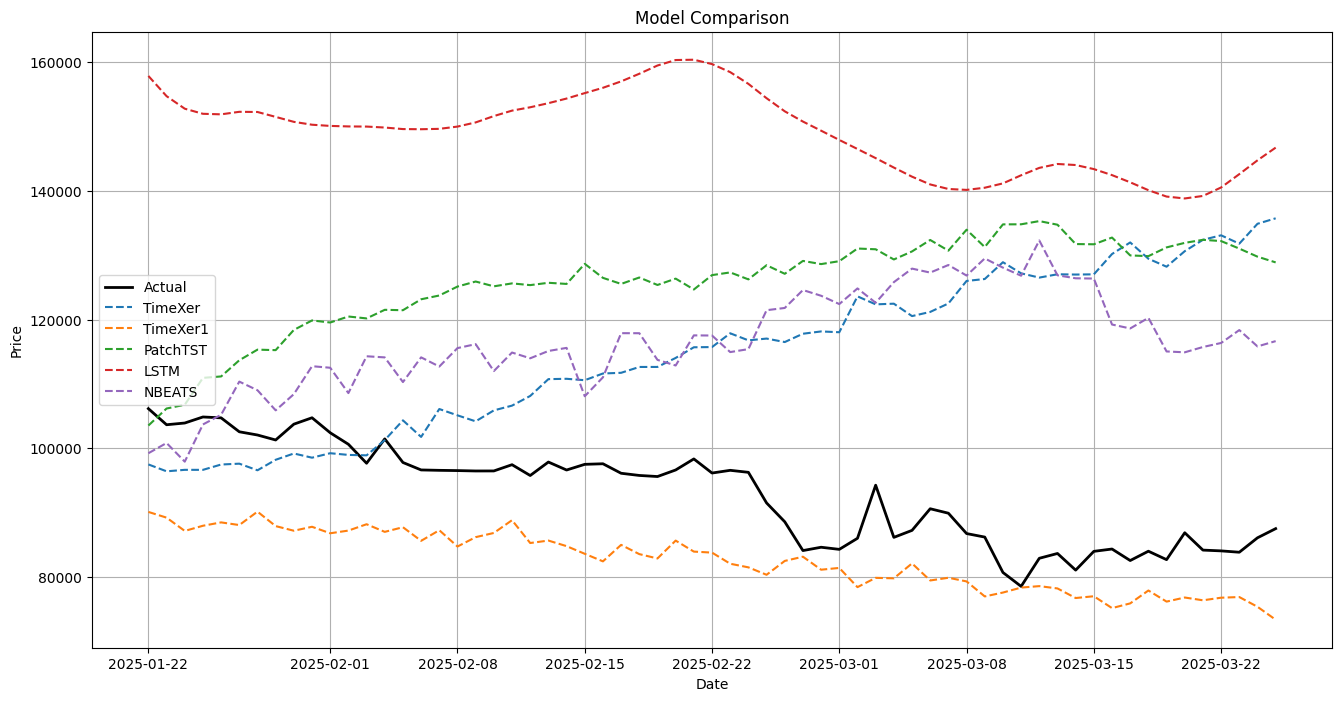}
\caption{Forecast comparison at horizon $H=63$ days.}
\label{fig:h63_forecast}
\end{figure}

 Figures~\ref{fig:h7_forecast}, \ref{fig:h28_forecast}, and \ref{fig:h63_forecast} offer a useful visual comparison between the model predictions, showing typical behavior at short, medium, and longer horizons.

At the shortest horizon ($H=7$), PatchTST tracks Bitcoin price most closely, as seen in Figure~\ref{fig:h7_forecast}. This outcome feels intuitive. Over a single week, price dynamics are largely shaped by very local patterns, recent momentum, and short-lived market reactions. In this setting, macroeconomic signals move slowly in the background and add little immediate guidance. As a result, the liquidity-aware TimeXer-Exog model does not yet separate itself from purely price-driven approaches, and any advantage from exogenous information doesn't help with the predictions.

At the medium horizon ($H=28$), the value of conditioning on macro-financial information becomes easier to see. In Figure~\ref{fig:h28_forecast}, the TimeXer-Exog forecast stays close to the actual Bitcoin price path, picking up both the sharp drawdown in March and the rebound that followed into May. Other models relying only on past prices tend to react more slowly, smoothing over the decline and lagging the recovery. The result is a loss of directional accuracy and a forecast that feels slightly out of sync with the actual price of Bitcoin.

By the time the horizon reaches $H=63$, the weaknesses of purely univariate models are obvious. In Figure~\ref{fig:h63_forecast}, many of the baseline forecasts begin to wander, drifting away from the realized price path and losing any clear sense of direction. Some flatten out, others veer off entirely, and in both cases the connection to the underlying market dynamics weakens. The TimeXer-Exog model behaves differently, its forecasts stay aligned with the broader trend, preserving directional consistency and a coherent structure over time. This shows the significance of the role of lagged global liquidity. Because global liquidity changes slowly but gives early signals, it helps long-term forecasts when price data alone is no longer sufficient.

\subsection{Interpretability Analysis: Visualizing Liquidity Transmission}

\begin{figure}[H]
    \centering
    \begin{minipage}{0.48\textwidth}
        \centering
        \includegraphics[width=\linewidth]{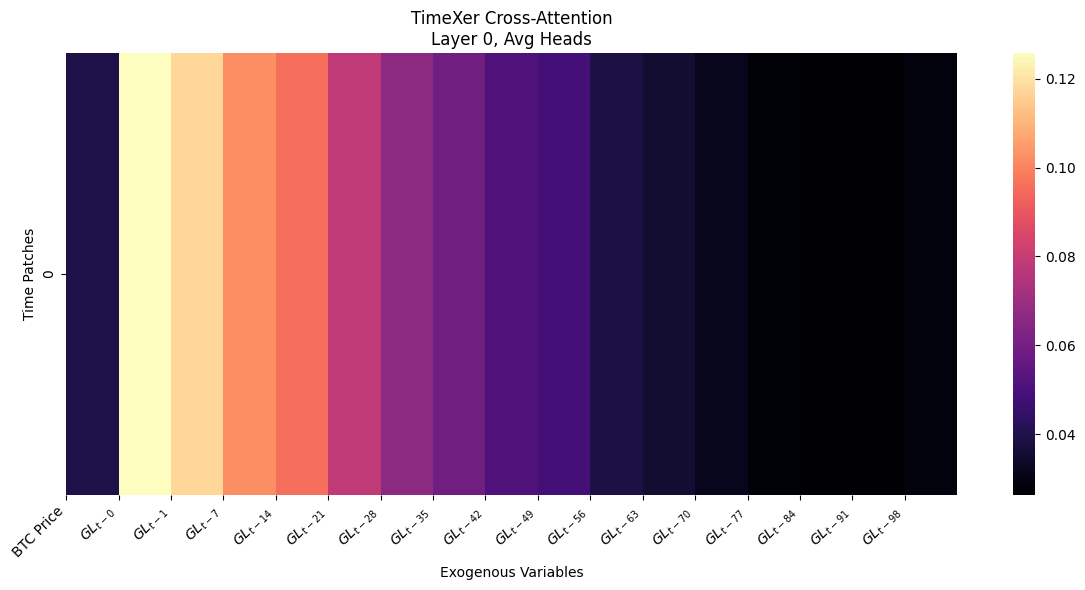}
        \subcaption{Horizon $H=7$ (Short)}
    \end{minipage}
    \hfill
    \begin{minipage}{0.48\textwidth}
        \centering
        \includegraphics[width=\linewidth]{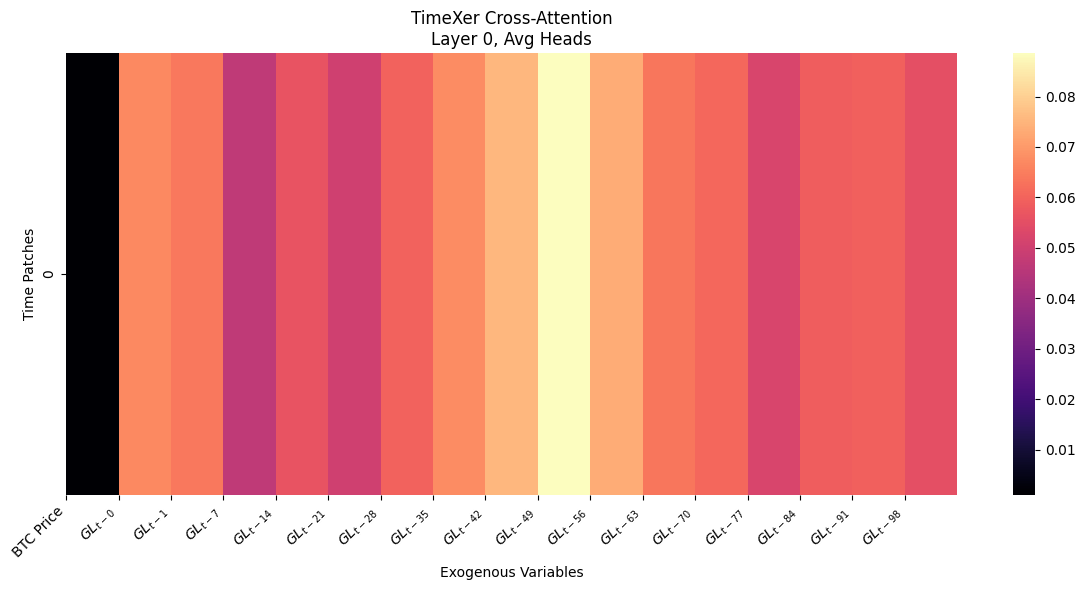}
        \subcaption{Horizon $H=21$ (Medium)}
    \end{minipage}
    
    \vspace{0.5cm}
    
    \begin{minipage}{0.6\textwidth}
        \centering
        \includegraphics[width=\linewidth]{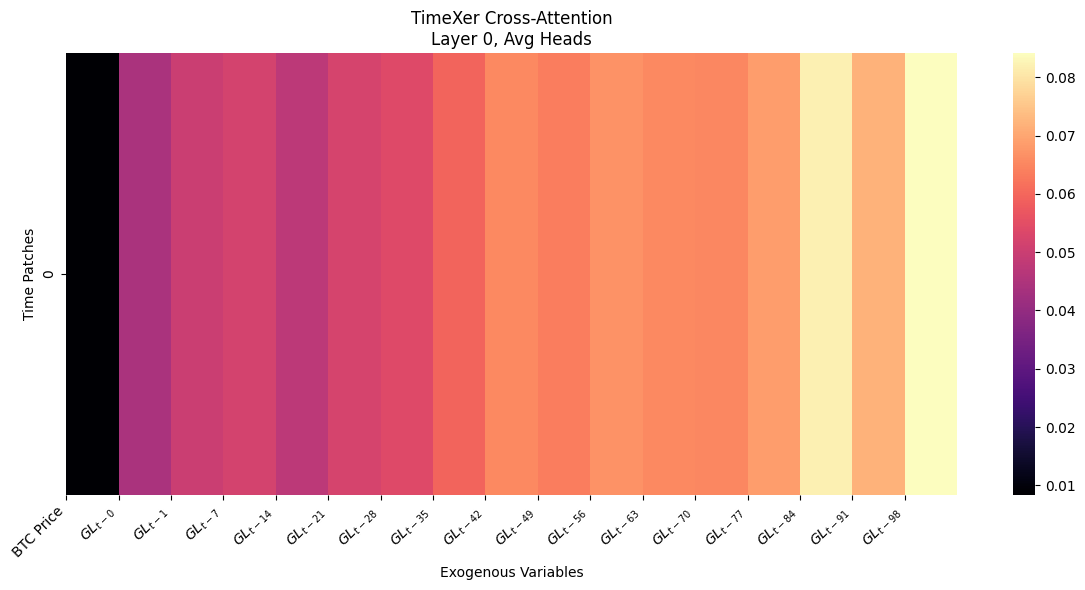}
        \subcaption{Horizon $H=63$ (Long)}
    \end{minipage}
    
    \caption{\textbf{Evolution of Cross-Attention Patterns (Layer 0).} (a) At $H=7$, attention is focused on immediate indicators (\texttt{global} and \texttt{global\_lag\_1}). (b) At $H=21$, a distinct anchor emerges at \texttt{global\_lag\_49} (7 weeks), suggesting a learned transmission delay. (c) At $H=63$, the focus shifts to deep lags (\texttt{global\_lag\_84} to \texttt{global\_lag\_91}), aligning with the quarterly liquidity cycle.}
    \label{fig:attention_maps}
\end{figure}

To validate whether the performance gains at long-horizons stem from global liquidity signals, we inspected the cross-attention weights from the model's first layer (Layer 0), averaged across all attention heads. Figure \ref{fig:attention_maps} visualizes which specific lags of Global Liquidity the model prioritizes when generating forecasts for different horizons.
While attention patterns evolve across layers, we find that Layer 0 offers the clearest and most stable representation of liquidity–price transmission. Visual inspection of deeper layers revealed qualitatively similar but increasingly diffuse patterns; hence, we report Layer 0 for interpretability and clarity.

The analysis reveals three distinct behavioral regimes:

\textbf{1. Short-Term Reactivity ($H=7$):}
At the shortest horizon (Figure \ref{fig:attention_maps}a), the model allocates the greatest attention to the most recent data points, particularly the target variable itself (\texttt{Bitcoin\_price}) and immediate liquidity measures (\texttt{global}, \texttt{global\_lag\_1}). This confirms that for weekly forecasts, the model operates primarily as a momentum trader, relying on current market conditions rather than delayed macroeconomic signals.

\textbf{2. Medium-Term Anchoring ($H=21$):}
A structural shift is observed at the 3-week horizon. As shown in Figure \ref{fig:attention_maps}(b), the attention mechanism shifts away from current data and forms a distinct peak at \texttt{global\_lag\_49}. This indicates that for 21-day predictions, the model autonomously learned to "look back" approximately 7 weeks. This 49-day lag likely captures the intermediate transmission period as liquidity changes filter through traditional financial systems before impacting risk assets.

\textbf{3. Long-Term Alignment ($H=63$):}
At the longest horizons (Figure \ref{fig:attention_maps}c), the attention weights shift to the deepest part of the context window, showing sustained focus on \texttt{global\_lag\_84} and \texttt{global\_lag\_91}. This corresponds almost exactly to a 12-week (3-month) cycle. The model's spontaneous reliance on these deep lags provides empirical support for the hypothesis that Global Liquidity acts as a leading indicator with a quarterly transmission lag, a relationship that simple univariate models fail to capture.

\subsection{Statistical Significance and Robustness Analysis}

Mean squared error is a simpler way to rank models from best to worst. However, it only scratches the surface. Prices jump for reasons that have little to do with forecasting skill, and a tiny edge in error can just as easily be luck as insight. That uncertainty tends to get glossed over when models are ranked mechanically.

This is where the MCS becomes useful. Table \ref{tab:mcs_results} reports the corresponding p-values across different forecast horizons, which forces a pause before declaring winners. Within this setup, a p-value of 1.000 means a model survives every comparison and stays in the set of top performers. Lower values tell a less flattering story. Once the threshold drops below 0.10, the supposed edge starts to look fragile, the kind that might disappear if the sample were reshuffled or extended by a few months. At that point, exclusion is hard to avoid.

From this perspective, the procedure does not magically identify the “true” best model. Rather, it reins in overconfidence. It helps distinguish improvements that persist under repeated scrutiny from those that shine briefly, perhaps because a particular market phase happened to line up in their favor. That distinction matters, especially when forecasting models are expected to hold up beyond the specific slice of data that first made them look good.

\begin{table}[ht]
\centering
\caption{MCS P-Values across Forecasting Horizons. \textbf{Bold} values indicate inclusion in the 90\% confidence set ($\alpha=0.10$). Models with $p < 0.10$ are statistically significantly inferior, meaning they are unlikely to
be the best model.}
\label{tab:mcs_results}
\resizebox{\textwidth}{!}{%
\begin{tabular}{lccccc}
\hline
\textbf{Horizon ($h$)} & \textbf{TimeXer-Exog (Ours)} & \textbf{TimeXer} & \textbf{PatchTST} & \textbf{LSTM} & \textbf{N-BEATS} \\ \hline
7  & \textbf{0.288} & 0.092 & \textbf{1.000} & \textbf{0.288} & 0.092 \\
14 & \textbf{1.000} & \textbf{0.671} & \textbf{0.671} & \textbf{0.186} & \textbf{0.186} \\
21 & \textbf{0.694} & \textbf{1.000} & 0.065 & \textbf{0.202} & \textbf{0.204} \\
28 & \textbf{1.000} & \textbf{0.305} & 0.077 & \textbf{0.305} & \textbf{0.108} \\
35 & \textbf{1.000} & 0.075 & 0.009 & 0.055 & 0.000 \\
42 & \textbf{1.000} & \textbf{0.202} & 0.001 & \textbf{0.443} & \textbf{0.443} \\
49 & \textbf{1.000} & 0.071 & 0.063 & \textbf{0.311} & 0.063 \\
56 & \textbf{1.000} & 0.086 & 0.086 & 0.086 & 0.033 \\
63 & \textbf{1.000} & 0.004 & 0.000 & 0.000 & 0.003 \\
70 & \textbf{1.000} & 0.005 & 0.000 & 0.037 & 0.000 \\ \hline
\end{tabular}%
}
\end{table}

\textbf{Short-Term Dynamics ($h \in [7, 21]$):}
Over shorter horizons, the MCS results point to a fairly crowded field. At $h=14$ and $h=21$, several models remain statistically indistinguishable, with the confidence set including {TimeXer-Exog, TimeXer, LSTM}. In practical terms, this suggests that when looking only a few weeks ahead, Bitcoin prices still carry a lot of their own momentum. Recent price history does much of the work. Well-designed univariate models, such as PatchTST, LSTM and NBEATS, can extract enough structure from autocorrelation alone. In these windows, adding global liquidity does not help much.

\textbf{Long-Term Superiority ($h \in [56, 70]$):}
Things change as the horizon increases. For forecasts at $h=63$ and $h=70$, the results in Table \ref{tab:mcs_results} are unambiguous. TimeXer-Exog is the only one in the confidence set, with a p-value of 1.000. All other models, including the base TimeXer and PatchTST, fall out well before the 10 percent threshold. This strongly suggests that incorporating exogenous information is very advantageous.

\textbf{Discussion of Results:}
This separation at longer horizons points directly to the role of global M2 liquidity. The relevance of past price movements fades with an increase in forecast horizons. As autoregressive signals begin to weaken, univariate models tend to lose stability, drifting toward long-run averages or accumulating forecast errors over time. Drawing on a 12-week lagged liquidity signal, the TimeXer-Exog model retains access to information that is not present in recent prices. Rather than reacting dramatically to the movement of the market, this signal leads it, offering a more stable reference point when price-based cues weaken. The MCS results suggest that the improvement in long-horizon accuracy is structural rather than accidental. In other words, the gains appear to come from how macroeconomic information is built into the model, not from statistical luck.

\section{Conclusion}

The results demonstrate that integrating Global M2 Liquidity into the TimeXer architecture evolves the model from a passive statistical tool into a robust \textbf{expert system for digital asset management}. Over forecast horizons extending to 70 days, the system’s ability to anchor predictions to macroeconomic signals yields a reduction in forecast error of over 89\% relative to univariate benchmarks. This performance gap confirms that the model successfully mimics expert intuition: it learns to prioritize slow-moving, high-impact monetary signals over short-term price noise when projecting long-term trends.

This framework serves as a critical \textbf{decision-support tool} for risk management and algorithmic trading. By explicating the relationship between global liquidity and asset prices through attention mechanisms, the system offers interpretability that purely technical models lack. It provides risk managers with ``early warning'' capabilities, preventing the model from overreacting to idiosyncratic price shocks while ensuring alignment with broader economic cycles. This capability allows for the construction of trading pipelines that reflect how sophisticated financial decisions are actually made—with one eye on market microstructure and the other on the global macroeconomic landscape.

Past prices are important, but without explicit links to macroeconomic drivers, forecasts tend to drift. Long-horizon forecasting in crypto markets or any market seems unlikely to work if it relies only on past prices. Attention-based architectures that incorporate exogenous liquidity signals offer a better way to connect predictions to underlying economic forces, which in turn makes them more resilient in highly volatile markets.

Future work could bring in more macro-financial variables, experiment with lag structures that adapt across different market regimes. We could test whether liquidity-conditioned forecasting transfers well to other cryptocurrencies and cross-asset portfolios. This could make forecasting systems reflect how real financial decisions are made, with one eye on prices and the other on the broader economic landscape, rather than just a technical tool.

\section*{Data Availability}

All data used here are publicly accessible. The Bitcoin prices, along with the processed global liquidity series, can be found in the GitHub repository at \url{https://github.com/sravankarthik/bitcoin-timeXer/blob/main/bitcoin.csv}
. The same repository also contains the full notebook needed to replicate the analysis, including the steps used to construct the global liquidity measure, introduce lagged features, and train all the models. The TradingView Pine Script for bitcoin and global liquidity is included as well \url{https://github.com/sravankarthik/bitcoin-timeXer}
. With these materials, the experiments reported in this paper should be reproducible in full, or at least close enough.

\bibliographystyle{apalike}
\bibliography{sample}

\end{document}